\documentclass{ecai} 

\usepackage{latexsym}
\usepackage{amssymb}
\usepackage{amsmath}
\usepackage{amsthm}
\usepackage{booktabs}
\usepackage{enumitem}
\usepackage{graphicx}
\usepackage{color}
\usepackage{pythonhighlight}
\usepackage{booktabs}
\usepackage{multirow}
\usepackage{makecell}
\usepackage{pifont}
\usepackage{siunitx}
\usepackage[caption=false]{subfig}
\usepackage{colortbl}
\usepackage{epsfig} 

\newcommand{\BibTeX}{B\kern-.05em{\sc i\kern-.025em b}\kern-.08em\TeX}
\newcommand{\methodname}{PROM }
\newcommand{\nospacemethodname}{PROM}
\newcommand{\cin}{C_{in}}
\newcommand{\cout}{C_{out}}
\newcommand{\plaincmark}{\ding{51}}%
\newcommand{\plainxmark}{\ding{55}}%
\definecolor{LightGray}{gray}{0.9}

\newif\ifarxiv

\arxivtrue 

\begin{document}


\begin{frontmatter}


\paperid{5602} 


\title{PROM: Prioritize Reduction of Multiplications Over Lower Bit-Widths for Efficient CNNs}


\author[A,B]{\fnms{Lukas}~\snm{Meiner}\thanks{Corresponding Author. Email: firstname.lastname@bosch.com}}
\author[A]{\fnms{Jens}~\snm{Mehnert}}
\author[A,B]{\fnms{Alexandru Paul}~\snm{Condurache}}

\address[A]{Robert Bosch GmbH, Leonberg, Germany}
\address[B]{Universit\"at zu L\"ubeck, L\"ubeck, Germany}


\begin{abstract}
	Convolutional neural networks (CNNs) are crucial for computer vision tasks on resource-constrained devices. Quantization effectively compresses these models, reducing storage size and energy cost. 
	However, in modern depthwise-separable architectures, the computational cost is distributed unevenly across its components, with pointwise operations being the most expensive. By applying a general quantization scheme to this imbalanced cost distribution, existing quantization approaches fail to fully exploit potential efficiency gains.
	To this end, we introduce \nospacemethodname, a straightforward approach for quantizing modern depthwise-separable convolutional networks by selectively using two distinct bit-widths. Specifically, pointwise convolutions are quantized to ternary weights, while the remaining modules use 8-bit weights, which is achieved through a simple quantization-aware training procedure. Additionally, by quantizing activations to 8-bit, our method transforms pointwise convolutions with ternary weights into int8 additions, which enjoy broad support across hardware platforms and effectively eliminates the need for expensive multiplications. 
	Applying \methodname to MobileNetV2 reduces the model's energy cost by more than an order of magnitude (23.9$\times$) and its storage size by 2.7$\times$ compared to the float16 baseline while retaining similar classification performance on ImageNet. Our method advances the Pareto frontier for energy consumption vs.\ top-1 accuracy for quantized convolutional models on ImageNet. 
	\methodname addresses the challenges of quantizing depthwise-separable convolutional networks to both ternary and 8-bit weights, offering a simple way to reduce energy cost and storage size.
\end{abstract}

\end{frontmatter}


\section{Introduction}
\label{sec:intro}
While computer vision models have made remarkable progress in the last decade \cite{He2016Deep,Howard2017MobileNets,Sandler2018MobileNetV2,Tan2021EfficientNetV2}, their increasing computational cost raises concerns about energy consumption, environmental impact, and suitability for deployment on resource-constrained devices. This is particularly challenging for large-scale Transformer models \cite{Dosovitskiy2021Image,Touvron2021Training}.
In contrast, convolutional neural networks (CNNs) remain widely used, especially in mobile environments, as they strike a balance between performance, resource requirements and training efficiency.

\begin{figure}[tbp]
	\centering
	\includegraphics[width=0.85\columnwidth]{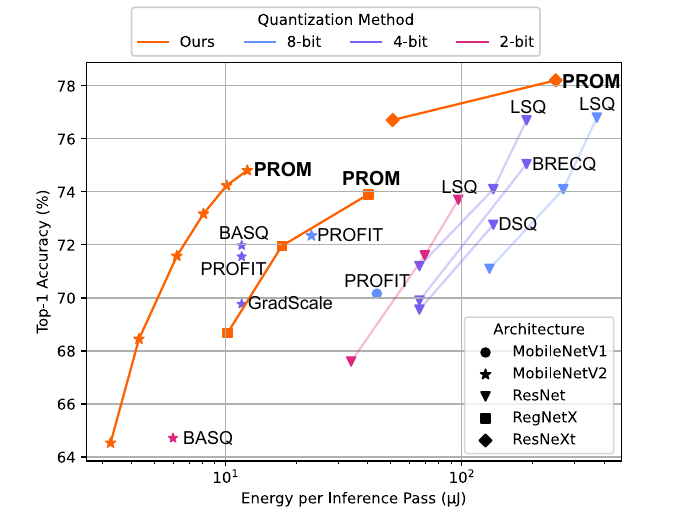}
	\caption{Comparison of quantized CNNs on ImageNet in terms of the trade-off between accuracy and energy consumption per forward pass in microjoules. The shape of a marker represents the underlying model architecture, while its color represents the quantization method used.}
	\label{fig:energy_pareto}
\end{figure}

While modern depthwise-separable CNNs offer an excellent trade-off between accuracy and efficiency, even these lightweight models can still be taxing for devices with limited compute resources. In particular, their most expensive components, namely $1\times1$ pointwise convolutions, dominate both parameter count and energy consumption, yet remain in full precision by default. To mitigate this, quantization can reduce floating-point weights and activations to lower bit-width integer formats, reducing both model size and computational cost. Quantization to 8-bit often preserves accuracy almost fully \cite{Park2020PROFIT,Zhu2020DSGC,Esser2020LSQ} and is widely supported and strongly optimized on general-purpose hardware \cite{IntelIntrinsics,ARMNEON,armIntroduction}. Dropping below 8-bit quantization quickly degrades task performance \cite{Gong2019Differentiable,Esser2020LSQ}, and 4-bit or 2-bit schemes typically require multi-stage training procedures \cite{Gong2019Differentiable,Esser2020LSQ,Li2021BRECQ,Park2020PROFIT} that progressively lower the bit-width, knowledge distillation \cite{Hinton2015Distilling} from a real-valued teacher \cite{Esser2020LSQ,Li2021BRECQ,Park2020PROFIT} or custom-built architectures \cite{Rastegari2016XNOR,Bethge2021MeliusNet}, and often lack native hardware support.

To address these issues and still push beyond int8 quantization, we introduce \nospacemethodname, a simple and novel approach to quantize modern depthwise-separable CNN architectures to both ternary and 8-bit weights. We tailor our quantization scheme to the common block structure found in these models, employing ternary weights (or 1.58 bits, since $\log_2(3)\approx1.58$) for high-cost pointwise $1\times1$ convolutions and 8-bit weights for the relatively inexpensive depthwise convolutions. This mixed-precision recipe retains the hardware-friendliness of int8 computations, eliminates the expensive multiplications in the heaviest layers, and sidesteps the training complexity pitfalls of sub-8-bit approaches.

Our proposed method prioritizes ease-of-use in terms of training, architecture design and deployability, while maintaining strong performance. Experiments on the ImageNet benchmark dataset \cite{Russakovsky2015ImageNet} demonstrate that our \methodname approach can generate a model that performs similarly to the floating-point model, while using 2.7$\times$ less storage and 23.9$\times$ less energy per forward pass. In a broader context, the models trained with our method form a new Pareto frontier for the trade-off between model accuracy and energy consumption per forward pass, while offering a more streamlined training process compared to other methods. 

Our main contributions can be summarized as follows:
\begin{itemize}
	\item We identify two characteristic features of the distribution of computational cost in depthwise-separable CNN architectures, and discuss how this negatively impacts the efficiency of existing quantization methods.
	\item Based on these findings, we introduce \nospacemethodname, a simple and effective method for quantizing depthwise-separable convolution models to ternary and 8-bit weights, enabling the use of heavily-optimized int8 addition routines.
	\item Our method achieves a considerable 23.9$\times$ reduction in the energy cost per forward pass of a MobileNetV2 architecture and reduces its storage size by 2.7$\times$, while retaining the performance of the real-valued model.
\end{itemize}

\section{Cost Analysis}
\label{sec:cost}
Depthwise-separable convolutions \cite{Chollet2017Xception,Howard2017MobileNets} were introduced as a parameter-efficient and computationally inexpensive alternative to dense convolutions with larger kernel sizes. Their basic structure is now used in most popular CNN architectures \cite{Howard2017MobileNets, Sandler2018MobileNetV2, Radosavovic2020Designing, Xie2017Aggregated} and consists of three major parts, as visualized in Figure~\ref{fig:architecture}a: 1) A pointwise convolution, functioning as an up-projection into a higher-dimensional latent space, 2) a depthwise convolution, extracting information independently from each channel using a $3\times3$ kernel, and 3) another pointwise convolution which projects the latent dimension into a lower output dimension.
Architectures like RegNet \cite{Radosavovic2020Designing} or ResNeXt \cite{Xie2017Aggregated} employ a similar structure, but use group sizes larger than one in their depthwise convolution blocks. 

\begin{figure}
	\centering
	\includegraphics[width=0.85\linewidth]{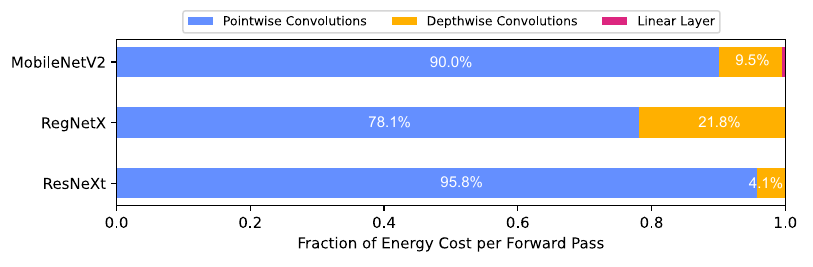}
	\caption{The distribution of energy cost per forward pass for different model architectures. Pointwise convolutions dominate the energy consumption, while depthwise convolutions and the linear layer require very little energy.}
	\label{fig:energy_split}
\end{figure}

\begin{figure}
	\centering
	\includegraphics[width=0.78\linewidth]{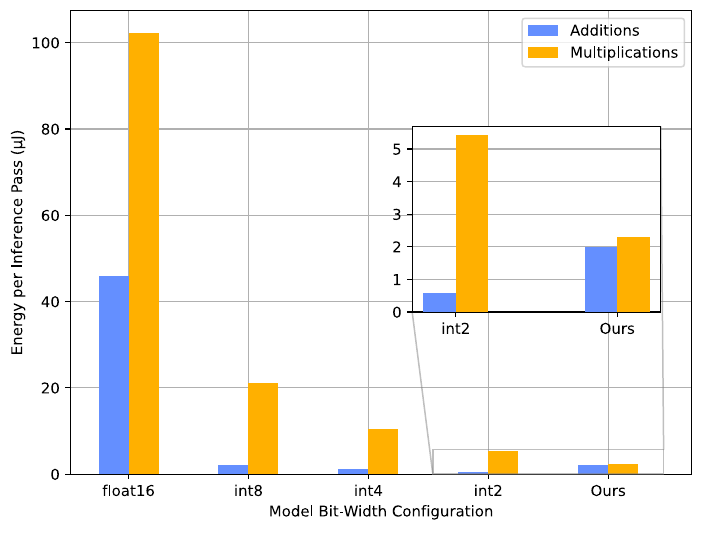}
	\caption{The energy consumption per inference pass of a MobileNetV2 model, quantized to different bit-widths (int8, int4, int2 and our proposed method), with the float16 model as a baseline for comparison. We visualize the cost of additions separately from the cost of multiplications. We find that multiplications consume the most energy by far, which is especially notable for the int8, int4 and int2 models.}
	\label{fig:energy_per_quant_method}
\end{figure}

\textbf{Cost is Shifted Towards Pointwise Convolutions.} For all of these models, the cost between different operations is not evenly distributed. Taking the MobileNetV2 model as an example, the depthwise convolutions only account for 1.9\% of the model's parameters and 9.5\% of the energy required during the forward pass, while all pointwise convolutions make up 61.2\% of the model's parameters and 90.0\% of its energy cost, excluding elementwise operations such as BatchNorm \cite{Ioffe2015Batch}. The remaining cost is mainly attributed to the fully connected linear layer. This observation is visualized in Figure~\ref{fig:energy_split}, where RegNet and ResNeXt models show similar distributions.

\textbf{Cost is Shifted Towards Multiplication Operations.} While it is no surprise that multiplications are more expensive to run on hardware than additions \cite{Horowitz2014Computings,Zhang2022PokeBNN}, the extent to which the cost of multiplications exceeds the cost of additions, especially for models quantized to integer weights, is notable. In particular, for a MobileNetV2 architecture quantized to int2, multiplications in a forward pass consume 9.5$\times$ more energy than all additions combined. 

\textbf{Impact on Quantization Efficiency.} 
The above analysis indicates that the bulk of a depthwise-separable model's cost lies in multiplications from pointwise convolutions. Hence, quantizing every convolution module to the same bit-width imposes a disproportionately strong restriction on the expressivity of groupwise convolution's weights in relation to the energy cost savings.
Therefore, it can be more beneficial to reduce the number of multiplications in a network compared to reducing its bit-width, as depicted in Figure~\ref{fig:energy_per_quant_method}. In fact, taking a MobileNetV2 model which is quantized to int8 format and "removing" multiplications from pointwise convolutions would lead to a larger efficiency gain than quantizing the entire network to 2-bit. 

To alleviate this issue, we propose to heavily quantize the costly pointwise projections of the model to ternary weights and allow other parts to remain at a higher bit-width, namely 8-bit. Firstly, this acknowledges the distribution shift of cost towards pointwise convolutions and reserves model capacity in places where it is cheap. Secondly, it enables us to eliminate all multiplications from pointwise convolution modules, as multiplication with ternary weights $\{-1,0,1\}$ reduces to a sum of input channels.
\section{Method}
\label{sec:method}
In this section, we introduce the quantization scheme used by \methodname and its specific application to the typical block structure used in depthwise-separable networks. Additionally, we outline the training process of our proposed method. 

\subsection{Quantization Scheme}
\label{subsec:quantization_scheme}
\begin{figure*}[htbp]
	\centering
	\subfloat[Vanilla depthwise-separable block structure.]{\includegraphics[width=0.33\linewidth]{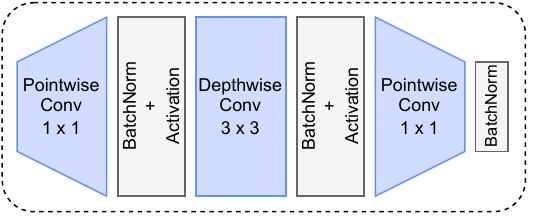}}
	\hspace{10mm}
	\subfloat[Our quantized block structure.]{\includegraphics[width=0.33\linewidth]{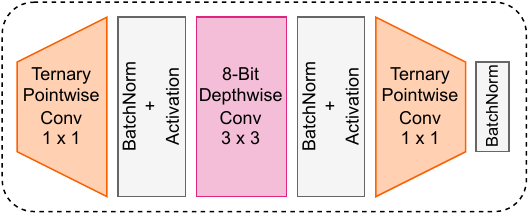}}
	\caption{Comparison between a regular depthwise-separable block structure and our quantized version. We quantize pointwise convolutions to ternary weights using channel-wise \textit{absmean} quantization. Depthwise convolutions are quantized to 8-bit integers using channel-wise \textit{absmax} quantization. This allows for accurate computations using a higher bit-width in between the efficient ternary convolutions.}
	\label{fig:architecture}
\end{figure*}
The quantization functions used in our proposed mixed quantization scheme are derived from the BitNet b1.58 \cite{Ma2024Era} language model, as they have demonstrated strong results while being simple to use. In particular, we employ channel-wise ternary \textit{absmean} quantization for all pointwise convolutions, channel-wise 8-bit \textit{absmax} quantization for all other convolutions and the linear layer, and tensor-wise 8-bit \textit{absmax} quantization for the activations.

\textbf{Pointwise Convolutions.} Let $W \in \mathbb{R}^{\cout \times \cin \times K \times K}$ be the weight matrix of a pointwise convolution layer, where $\cout$ and $\cin$ are the output and input channel dimensions, respectively. Since the kernel size $K=1$, we can omit the added dimensions and write $W \in \mathbb{R}^{\cout \times \cin}$. To quantize the layer to ternary weights, we first compute the mean absolute value per output channel as a scale factor:
\begin{equation}
	\alpha_i = \frac{1}{\cin} \sum_{j=1}^{\cin} |W_{i,j}|, \quad \alpha \in \mathbb{R}^{\cout}.
	\label{eq:scale_factor_alpha}
\end{equation}
Using this scale, we generate the quantized weights $\hat{W} \in \{-1,0,1\}^{\cout \times \cin}$ by rounding and clamping:
\begin{equation}
	\text{RoundClip}(x,a,b) = \max(a,\,\min(b,\,\text{round}(x))),
\end{equation}
\begin{equation}
	\hat{W}_i = \text{RoundClip}\!\left(\frac{W_i}{\alpha_i + \epsilon}, -1, 1\right),
\end{equation}
where we use $\epsilon = 10^{-5}$ to avoid division by zero. 
During the quantization-aware model training, the layer's forward pass is computed by dequantizing each output channel $\hat{W}_i$ through multiplying it with $\alpha_i$ before convolving with the input to ensure proper gradient calculation. At inference time, the input is directly convolved with the ternary weights $\hat{W}$, reducing the convolution to a sum of input values, and scaled by $\alpha$ afterwards.

\textbf{Depthwise Convolutions and Linear Layer.} Depthwise convolution layers can be represented by a learned weight matrix $W \in \mathbb{R}^{\cout \times \cin \times K \times K}$ with kernel size $K>1$. We compute the channel-wise scale as the \textit{absmax} of their weights:
\begin{equation}
	\beta_i = \max_{j,k,l} |W_{i,j,k,l}|, \quad \beta \in \mathbb{R}^{\cout}
\end{equation}
and quantize the weights to 8-bit precision:
\begin{equation}
	\hat{W}_{i,j,k,l} = \text{RoundClip}\!\left(\frac{W_{i,j,k,l}}{\beta_i + \epsilon}, -128, 127\right).
\end{equation}
The same quantization scheme is applied to the model's linear layer, where we determine the \text{absmax} of the weight values across all input channels.

\textbf{Activations.}
To keep computational cost low during inference, we choose tensor-wise quantization for the activations of the model and use the same \textit{absmax} scheme as before. For a given input $X \in \mathbb{R}^{B \times \cin \times H \times W}$ to a layer, where $B$ is the batch size and $H$ and $W$ are the inputs height and width, respectively, we determine the quantization scale factor per batch element instead of per channel:
\begin{equation}
	\gamma_i = \max_{j,k,l} |X_{i,j,k,l}|, \quad \gamma \in \mathbb{R}^B
\end{equation}   
and quantize the activations to 8-bit:
\begin{equation}
	\hat{X}_{i,j,k,l} = \text{RoundClip}\!\left(\frac{X_{i,j,k,l}}{\gamma_i + \epsilon}, -128, 127\right).
\end{equation}

\subsection{Training Process and Adaptations}
\label{subsec:TrainingProcessAndAdaptations}
To train a model with our proposed quantization scheme, we use standard quantization-aware training techniques and closely follow the training routine of the vanilla model as found in the TorchVision \cite{maintainers2016TorchVision} training recipe. During training, the model's underlying weights remain in 32-bit floating-point precision and can be updated using standard gradient descent. They are quantized and dequantized on the fly and can later be converted to fixed quantized weights for efficient inference.

To propagate gradients through the round function, we follow common practice and employ the straight-through estimator (STE) \cite{Hubara2016Binarized}. This allows us to use standard optimization algorithms such as stochastic gradient descent (SGD).

As a baseline, we train the vanilla MobileNetV2 \cite{Sandler2018MobileNetV2}, RegNetX \cite{Radosavovic2020Designing} and ResNeXt \cite{Xie2017Aggregated} models using the hyperparameters presented in their respective TorchVision recipe.
For the quantized models, we use the same hyperparameter choices and only make one modification: We replace the learning rate scheduler by a cosine decay strategy, which lowers the learning rate close to zero towards the end of training. This has been found to aid convergence of models using binary or ternary quantization \cite{Wang2023BitNet,Zhang2023Binarized,Ma2024Era,Zhu2024Scalable}.

For the MobileNetV2 and RegNetX architectures, we employ two additional modifications that increase the quantized model's performance. Firstly, we set the weight decay to zero halfway through the training process to aid convergence. Its use in the training of ternary weights differs from that in full-precision training, with non-zero weight decay values encouraging fluctuation in the ternary weights \cite{Ma2024EraSupp}.
Secondly, we replace all ReLU6 or ReLU activations in these two architectures with PReLU \cite{Redfern2021BCNN,Martinez2020Real2Bin,Guo2023BNext,Zhang2022PokeBNN,Liu2020ReActNet}. This is a simple and computationally inexpensive way to restore some of the expressivity of the model which is lost in the quantization process.
A detailed component ablation of these choices is found in Section~\ref{sec:ablation_study}.
\section{Evaluating Model Efficiency}
We evaluate our proposed quantization scheme against related methods in terms of energy consumption, model size, and hardware support.
Energy consumption is a key indicator of usability in battery-powered devices and strongly correlates with different inference cost metrics \cite{Zhang2022PokeBNN}. Lower memory consumption ensures broad applicability and positively affects load times of the model on resource-constrained and mobile devices, reducing computational overhead. Lastly, hardware support for quantization operations is essential for the usability and reproducibility of a method. By evaluating our approach in these areas, we aim to provide a comprehensive assessment of its impact and applicability to real-world scenarios. 

\textbf{Energy.}
Throughout the paper, we follow prior work \cite{Zhang2022PokeBNN, Guo2023BNext} and estimate the total energy cost for an inference pass of a model using operation-level energy consumption measurement tables provided in \cite{Horowitz2014Computings, Zhang2022PokeBNN}. They give an overview of the energy consumption of ADD and MUL operations for float32, float16, int32 and int8 data types on \SI{45}{\nano\meter} and \SI{7}{\nano\meter} process nodes, measured in femtojoules (\unit{\femto\joule}). By determining the exact number of operations for each data type, we can estimate the overall energy requirement of a model. 

As energy measurements can vary drastically between different hardware (CPUs, GPUs or specialized accelerators) and their respective instruction sets, using a common reference \cite{Horowitz2014Computings, Zhang2022PokeBNN} ensures a fair comparison between different quantization approaches. By convention, we omit data transfer overhead in the energy analysis. However, since memory transfer typically scales proportionately with a reduction in model size, our \methodname approach likely stands to gain efficiency in comparison to the baseline if we factor in memory-access costs, measured in \SI{}{\joule} per 32-bit read operation. 

To be able to compare to int4 and int2 quantization methods as well, we make the assumption that the energy cost halves when going from int8 to int4, and from int4 to int2 computations, similar to \cite{Zhang2022PokeBNN}. By ignoring the real-world cost overhead of packing and unpacking int2 or int4 operands into larger registers found in commodity hardware, this assumption underestimates the energy use of sub-8-bit methods. Consequently, the advantage of \methodname over these approaches with respect to energy savings is a conservative estimate and likely improves in real-world scenarios.

We follow prior work \cite{Zhang2022PokeBNN,Guo2023BNext,Bethge2021MeliusNet,Liu2020ReActNet,Liu2018BiReal} and do not include cheap elementwise operations, such as batch normalization \cite{Ioffe2015Batch}, in the energy calculation for a fair comparison across methods.

\textbf{Memory.}
We report a model's storage size in megabytes (\unit{\mega\byte}). For other works, we either use their reported storage size if available, or compute it based on the bit-widths for each component. If floating-point values are used, we assume that the same accuracy could be achieved by using float16 instead of float32 formats, similar to \cite{Zhang2022PokeBNN}.

\textbf{Hardware Support.}
We discuss the availability of operations required to run different quantization schemes on commodity hardware. While methods relying on int8 computations are readily supported on modern CPUs and GPUs, lower bit-widths are harder to use without specific accelerator chips. 
Similarly, mixed-precision approaches \cite{Chen2021Towards,Chauhan2023Post,Yang2021FracBits,Wang2020Hardware} that learn an independent bit-width for each module in the network complicate hardware deployment, as they do not rely on a fixed quantization format across layers.
In contrast, \methodname uses precisely ternary weights for pointwise convolutions and 8-bit weights for depthwise convolutions as well as linear layers, simplifying the hardware instructions required for running inference with a model. 
\section{Experiments}
\label{sec:experiments}
In this section, we extensively evaluate our method and provide a comparison to related work based on results on the standard ImageNet ILSVRC 2012 benchmark \cite{Russakovsky2015ImageNet} for image classification. We follow common practice and report our results on the validation set of ImageNet. All of our models are implemented in PyTorch \cite{Paszke2019PyTorch} and trained as described in Section~\ref{subsec:TrainingProcessAndAdaptations}. 
\ifarxiv
	Additionally, we include pseudocode for our implementation in the Supplementary Material (see Figures~\ref{fig:code_quantization_of_conv_weights}-\ref{fig:code_quantized_conv_module}).
\else
	Additionally, we provide pseudocode for our implementation in the Supplementary Material (see Figures~10-12 in \cite{Meiner2025PROM}).
\fi
For the MobileNetV2 architecture \cite{Sandler2018MobileNetV2}, we train a suite of models by altering the width multiplier setting, ranging from 0.75$\times$ to 2.0$\times$. This setting adjusts the width of each module by the given factor. Similarly, we train RegNetX \cite{Radosavovic2020Designing} and ResNeXt \cite{Xie2017Aggregated} models with varying depths, allowing us to observe the scaling properties of our method.

\subsection{Results on ImageNet}
\label{subsec:results_on_imagenet}
\begin{figure*}[htbp]
	\centering
	\subfloat[Top-1 Accuracy vs. \SI{7}{\nano\meter} Inference Energy.]{\includegraphics[width=0.4\linewidth]{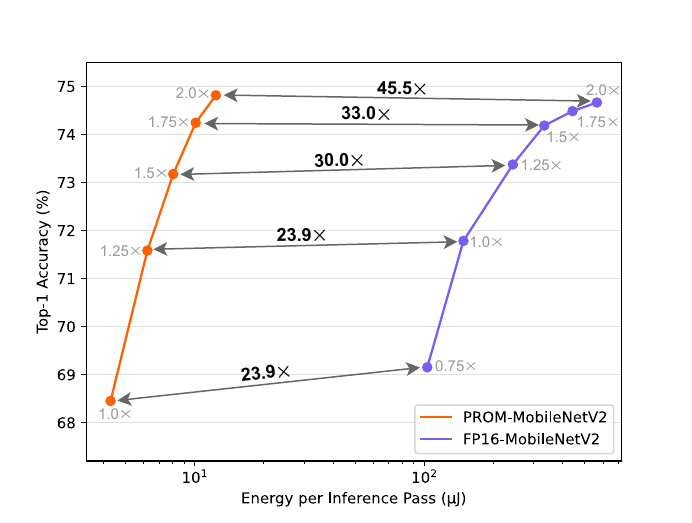}}
	\hspace{5mm}
	\subfloat[Top-1 Accuracy vs. Model Storage Size.]{\includegraphics[width=0.4\linewidth]{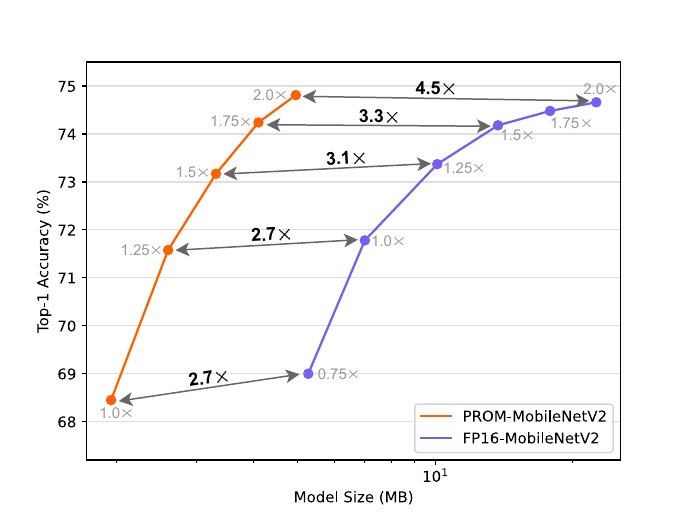}}
	\caption{Accuracy-resource trade-off for models quantized with \methodname compared to a float16 baseline MobileNetV2 architecture. Each curve shows Top-1 accuracy as we sweep the \textcolor{gray}{width multiplier} (denoted in \textcolor{gray}{gray}) for the model from 0.75$\times$ (float16 models) or 1.0 $\times$ (\methodname models) to 2.0$\times$. The labels above each arrow denote the factor by which \methodname reduces (a) energy consumption or (b) storage size relative to the float16 baseline.}
	\label{fig:energy_memory_comparison_to_baseline}
\end{figure*}

We present an overview of our results in comparison to related methods in Tables~\ref{tab:mobilenetv2_results}, \ref{tab:regnet_results} and \ref{tab:resnext_results} for the MobileNetV2, RegNetX and ResNeXt models, respectively. We compare results in terms of top-1 accuracy on ImageNet, model size in megabytes and energy consumption on \SI{45}{\nano\meter} and \SI{7}{\nano\meter} process nodes in microjoules. The energy consumption of our quantized models in comparison to other methods is visualized in Figure~\ref{fig:energy_pareto}. 

\begin{table}[htbp]
	\centering
	\caption{Comparison of results for MobileNetV2 models on the ImageNet benchmark. "W/A" denotes the bit-widths used for model weights and activations, respectively. 
		The results are categorized by similar accuracy. The best result per category is marked \textbf{bold}, the second best is \underline{underlined}.}
	\label{tab:mobilenetv2_results}
	\resizebox{1.0\linewidth}{!}{%
		\begin{tabular}{ccccrrr}
			\toprule
			\multirow[c]{2}{*}{Method} & \multirow[c]{2}{*}{Model} & \multirow[c]{2}{*}{W/A} & \multirow[c]{2}{*}{\makecell{Top-1 \\(\%)}} & \multirow[c]{2}{*}{\makecell{Size \\ (\unit{\mega\byte})}} & \multicolumn{2}{c}{Energy (\unit{\micro\joule})} \\
			& & & & & \SI{45}{\nano\meter} & \SI{7}{\nano\meter} \\
			\midrule
			\multirow[c]{6}{*}{Baseline} & 0.75$\times$MobileNetV2 & 16/16 & 69.15 & 5.27 & 308.7 & 102.6 \\
			& 1.0$\times$MobileNetV2 & 16/16 & 71.78 & 7.01 & 445.4 & 148.1 \\
			& 1.25$\times$MobileNetV2 & 16/16 & 73.37 & 10.10 & 733.0 & 242.8 \\
			& 1.5$\times$MobileNetV2 & 16/16 & 74.18 & 13.72 & 1000.4 & 332.9 \\
			& 1.75$\times$MobileNetV2 & 16/16 & 74.48 & 17.84 & 1326.7 & 441.5 \\
			& 2.0$\times$MobileNetV2 & 16/16 & 74.66 & 22.52 & 1694.5 & 564.1 \\
			\midrule
			BASQ \cite{Kim2022BASQ} & 1.0$\times$MobileNetV2 & 2/2 & \underline{64.71} & \textbf{0.94} & 17.3 & 6.0 \\
			\rowcolor{LightGray}
			\methodname & 0.75$\times$MobileNetV2 & (1.58/8)/8 & 64.53 & \underline{1.70} & \textbf{11.3} & \textbf{3.3} \\ 
			\rowcolor{LightGray}
			\methodname & 1.0$\times$MobileNetV2 & (1.58/8)/8 & \textbf{68.45} & 1.95 & \underline{15.2} & \underline{4.3} \\ 
			\midrule
			GradScale \cite{Sun2020GradScale} & 1.0$\times$MobileNetV2 & 4/4 & 69.77 & \textbf{1.80} & 34.4 & 11.7 \\
			PROFIT \cite{Park2020PROFIT} & 1.0$\times$MobileNetV2 & 4/4 & 71.56 & \textbf{1.80} & 34.4 & 11.7 \\
			BASQ \cite{Kim2022BASQ} & 1.0$\times$MobileNetV2 & 4/4 & \underline{71.98} & \textbf{1.80} & 34.4 & 11.7 \\
			\rowcolor{LightGray}
			\methodname & 1.25$\times$MobileNetV2 & (1.58/8)/8 & 71.58 & \underline{2.60} & \textbf{24.5} & \textbf{6.2} \\
			\rowcolor{LightGray}
			\methodname & 1.5$\times$MobileNetV2 & (1.58/8)/8 & \textbf{73.17} & 3.31 & \underline{29.5} & \underline{8.1} \\
			\midrule
			PROFIT \cite{Park2020PROFIT} & 1.0$\times$MobileNetV2 & 8/8 & 72.35 & \underline{3.54} & 68.7 & 23.1 \\
			\rowcolor{LightGray}
			\methodname & 1.5$\times$MobileNetV2 & (1.58/8)/8 & 73.17 & \textbf{3.31} & \textbf{29.5} & \textbf{8.1} \\
			\rowcolor{LightGray}
			\methodname & 1.75$\times$MobileNetV2 & (1.58/8)/8 & \underline{74.24} & 4.10 & \underline{37.5} & \underline{10.1} \\		
			\rowcolor{LightGray}
			\methodname & 2.0$\times$MobileNetV2 & (1.58/8)/8 & \textbf{74.81} & 4.96 & 46.4 & 12.4 \\
			\bottomrule 
		\end{tabular}
	}
\end{table}

\begin{table}[htbp]
	\centering
	\caption{Comparison of results for RegNet models on the ImageNet benchmark. "W/A" denotes the bit-widths used for model weights and activations, respectively. 
		The results are categorized by similar accuracy. The best result per category is marked \textbf{bold}, the second best is \underline{underlined}.}
	\label{tab:regnet_results}
	\resizebox{1.0\linewidth}{!}{%
		\begin{tabular}{ccccrrr}
			\toprule
			\multirow[c]{2}{*}{Method} & \multirow[c]{2}{*}{Model} & \multirow[c]{2}{*}{W/A} & \multirow[c]{2}{*}{\makecell{Top-1 \\(\%)}} & \multirow[c]{2}{*}{\makecell{Size \\ (\unit{\mega\byte})}} & \multicolumn{2}{c}{Energy (\unit{\micro\joule})} \\
			& & & & & \SI{45}{\nano\meter} & \SI{7}{\nano\meter} \\
			\midrule
			\multirow[c]{3}{*}{Baseline} & RegNetX-400MF & 16/16 & 72.69 & 10.99 & 650.4 & 215.8 \\
			& RegNetX-800MF & 16/16 & 75.27 & 14.52 & 1308.8 & 434.7 \\
			& RegNetX-1.6GF & 16/16 & 76.90 & 18.38 & 2563.9 & 850.8 \\
			\midrule
			LSQ \cite{Esser2020LSQ} & ResNet18 & 2/2 & 67.60 & \underline{2.94} & \underline{97.2} & \underline{34.0} \\
			DSQ \cite{Gong2019Differentiable} & ResNet18 & 4/4 & \underline{69.56} & 10.18 & 193.3 & 66.2 \\
			BRECQ \cite{Li2021BRECQ} & ResNet18 & 4/4 & \textbf{69.90} & 5.81 & 193.3 & 66.2 \\
			\rowcolor{LightGray}
			\methodname & RegNetX-400MF & (1.58/8)/8 & 68.68 & \textbf{2.40} & \textbf{33.3} & \textbf{10.2} \\
			\midrule
			PROFIT \cite{Park2020PROFIT} & MobileNetV1 & 8/8 & 70.16 & \underline{4.25} & \underline{130.5} & \underline{43.7} \\
			LSQ \cite{Esser2020LSQ} & ResNet18 & 8/8 & 71.10 & 11.69 & 386.6 & 131.1 \\
			LSQ \cite{Esser2020LSQ} & ResNet18 & 4/4 & 71.20 & 5.85 & 193.3 & 66.2 \\
			LSQ \cite{Esser2020LSQ} & ResNet34 & 2/2 & \underline{71.60} & 5.48 & 199.6 & 69.7 \\
			\rowcolor{LightGray}
			\methodname & RegNetX-800MF & (1.58/8)/8 & \textbf{71.96} & \textbf{3.01} & \textbf{57.7} & \textbf{17.3} \\
			\midrule
			DSQ \cite{Gong2019Differentiable} & ResNet34 & 4/4 & 72.76 & 15.24 & 396.8 & 135.8 \\
			LSQ \cite{Esser2020LSQ} & ResNet50 & 2/2 & \underline{73.70} & \underline{6.48} & \underline{278.9}	& \underline{96.7} \\
			\rowcolor{LightGray}
			\methodname & RegNetX-1.6GF & (1.58/8)/8 & \textbf{73.91} & \textbf{4.15} & \textbf{131.1} & \textbf{40.3} \\
			\bottomrule 
		\end{tabular}
	}
\end{table}

\begin{table}[htbp]
	\centering
	\caption{Comparison of results for ResNeXt models on the ImageNet benchmark. "W/A" denotes the bit-widths used for model weights and activations, respectively. 
		The results are categorized by similar accuracy. The best result per category is marked \textbf{bold}, the second best is \underline{underlined}.}
	\label{tab:resnext_results}
	\resizebox{1.0\linewidth}{!}{%
		\begin{tabular}{ccccrrr}
			\toprule
			\multirow[c]{2}{*}{Method} & \multirow[c]{2}{*}{Model} & \multirow[c]{2}{*}{W/A} & \multirow[c]{2}{*}{\makecell{Top-1 \\(\%)}} & \multirow[c]{2}{*}{\makecell{Size \\ (\unit{\mega\byte})}} & \multicolumn{2}{c}{Energy (\unit{\micro\joule})} \\
			& & & & & \SI{45}{\nano\meter} & \SI{7}{\nano\meter} \\
			\midrule
			\multirow[c]{2}{*}{Baseline} & ResNeXt-50(32$\times$4d) & 16/16 & 77.42 & 50.04 & 7520.0 & 2504.3 \\
			& ResNeXt-101(32$\times$8d) & 16/16 & 79.18 & 177.56 & 25567.5 & 8504.9 \\
			\midrule
			LSQ \cite{Esser2020LSQ} & ResNet34 & 8/8 & 74.10 & 21.81 & 793.5 & 269.2 \\
			LSQ \cite{Esser2020LSQ} & ResNet34 & 4/4 & 74.10 & \underline{10.92} & \underline{396.8} & \underline{135.8} \\
			BRECQ \cite{Li2021BRECQ} & ResNet50 & 4/4  & 75.05 & 12.85 & 553.6 & 188.4 \\
			LSQ \cite{Esser2020LSQ} & ResNet50 & 4/4 & \underline{76.70} & 12.85 & 553.6 & 188.4 \\
			\rowcolor{LightGray}
			\methodname & ResNeXt-50(32$\times$4d) & (1.58/8)/8 & \textbf{76.71} & \textbf{8.97} & \textbf{195.1} & \textbf{51.0} \\
			\midrule
			LSQ \cite{Esser2020LSQ} & ResNet50 & 8/8 & \underline{76.80} & \textbf{25.60} & \underline{1107.3} & \underline{372.6} \\
			\rowcolor{LightGray}
			\methodname & ResNeXt-101(32$\times$8d) & (1.58/8)/8 & \textbf{78.21} & \underline{32.09} & \textbf{878.8} & \textbf{249.6} \\
			\bottomrule 
		\end{tabular}
	}
\end{table}

By scaling the quantized model's width (MobileNetV2) and depth (RegNetX and ResNeXt), we achieve accurate and highly efficient models for varying storage size and energy budgets. For example, we can reduce the energy consumption of a 1.0$\times$MobileNetV2 model by a factor of up to 34.4$\times$, and the required memory to store it by 3.6$\times$. To nearly match the accuracy of the floating-point 1.0$\times$MobileNetV2, our 1.25$\times$-scaled model requires 2.7$\times$ less storage and 23.9$\times$ less energy per forward pass, as depicted in Figure~\ref{fig:energy_memory_comparison_to_baseline}. Interestingly, our largest model surpasses the floating-point baseline model's accuracy by 3.03 percentage points, while using 1.4$\times$ less storage and 11.9$\times$ less energy on \SI{7}{\nano\meter} architectures.

We observe that our method scales well with the model's width or depth. Strong quantization in the form of ternary weights for pointwise convolutions heavily restricts the function approximation properties of the model. This issue can be alleviated by providing it with more weights to tune, increasing its representational power. 
We also note that the 2.0$\times$-scaled MobileNetV2 exhibits a much greater amount of ternary weights equal to zero after training when compared to the 1.0$\times$ model, as illustrated in 
\ifarxiv
	Figure~\ref{fig:distribution_of_zeros} of the Supplementary Material.
\else
	Figure~7 of the Supplementary Material (see \cite{Meiner2025PROM}).
\fi
We hypothesize that this indicates an automatic allocation of model capacity during training, essentially "pruning" filters in unimportant layers. As weights which are equal to zero do not need to be computed in the forward pass, the energy expense can be reduced even further on suitable hardware.

\begin{table}[htbp]
	\centering
	\caption{Comparison of our method with approaches that learn a "Mixed" bit-width assignment per module. "W/A" denotes the bit-widths used for model weights and activations, respectively. The best result per category is marked \textbf{bold}, the second best is \underline{underlined}.}
	\label{tab:mixed_precision_results}
	\resizebox{1.0\linewidth}{!}{%
		\begin{tabular}{ccccrrr}
			\toprule
			\multirow[c]{2}{*}{Method} & \multirow[c]{2}{*}{Model} & \multirow[c]{2}{*}{W/A} & \multirow[c]{2}{*}{\makecell{Top-1 \\(\%)}} & \multirow[c]{2}{*}{\makecell{Size \\ (\unit{\mega\byte})}} & \multicolumn{2}{c}{Energy (\unit{\micro\joule})} \\
			& & & & & \SI{45}{\nano\meter} & \SI{7}{\nano\meter} \\
			\midrule
			HAQ \cite{Wang2020Hardware} & 1.0$\times$MobileNetV2 & Mixed & 70.9 & 3.03 & 55.9 & 18.8 \\
			\citet{Chen2021Towards} & 1.0$\times$MobileNetV2 & Mixed & 71.2 & \textbf{1.06} & 31.6 & 10.6 \\
			FracBits \cite{Yang2021FracBits} & 1.0$\times$MobileNetV2 & Mixed & \underline{71.9} & \underline{2.30} & 28.9 & 9.7 \\
			\rowcolor{LightGray}
			PROM & 1.25$\times$MobileNetV2 & (1.58/8)/8 & 71.6 & 2.60 & \textbf{24.5} & \textbf{6.2} \\
			\rowcolor{LightGray}
			PROM & 1.5$\times$MobileNetV2 & (1.58/8)/8 & \textbf{73.2} & 3.31 & \underline{29.5} & \underline{8.1} \\
			\bottomrule 
		\end{tabular}
	}
\end{table}

We also compare our method to recent mixed-precision approaches \cite{Chen2021Towards,Chauhan2023Post,Yang2021FracBits,Wang2020Hardware} that actively learn the bit-width per module in Table~\ref{tab:mixed_precision_results}. The results demonstrate that our principled approach provides a more effective path to model efficiency. For example, our \nospacemethodname-1.25$\times$MobileNetV2 achieves comparable accuracy to the best-performing learned mixed-precision method, while using 36\% less energy on a \SI{7}{\nano\meter} architecture. Moreover, our \nospacemethodname-1.5$\times$MobileNetV2 model surpasses all related methods in accuracy while still maintaining superior energy efficiency. This demonstrates that our fixed, hardware-friendly scheme with ternary weights for costly pointwise convolutions and 8-bit weights for all other layers is a more practical and effective strategy for resource-constrained inference than complex, learned bit-width assignments. 

\subsection{Energy Evaluation}
As visualized in Figure~\ref{fig:energy_pareto}, our method improves the Pareto frontier in the trade-off between top-1 accuracy and inference energy cost for CNN models on ImageNet. In particular, \methodname outperforms other methods for quantizing a MobileNetV2 to lower bit-widths. 

Our method achieves best-in-category energy efficiency on \SI{45}{\nano\meter} and \SI{7}{\nano\meter} architectures while offering similar or better performance than prior work, as demonstrated in Tables~\ref{tab:mobilenetv2_results}, \ref{tab:regnet_results} and \ref{tab:resnext_results}. This makes our method well-suited for deployment on battery-powered or resource-constrained devices. Additionally, note that the energy consumption of a model strongly correlates with the ACE metric introduced in \cite{Zhang2022PokeBNN}, which measures a network's total inference cost on hardware.

\subsection{Memory Evaluation}
For the MobileNetV2, RegNetX and ResNeXt architectures, \methodname is able to reduce a model's storage size for a given parameter budget by a factor of up to 3.6$\times$, 4.8$\times$ and 5.6$\times$, respectively.
Since our method relies on using both ternary weights and 8-bit weights, the models trained with \methodname naturally require less storage than 8-bit models on the same architecture. In general, we observe that the use of ternary weights in the pointwise convolutions, which make up the majority of model parameters in all tested architectures, allow us to keep a competitively low memory requirement. 

We also find that 4-bit and 2-bit MobileNetV2 models require less storage size than our models, but use more energy per inference pass on both \SI{45}{\nano\meter} and \SI{7}{\nano\meter} architectures. For example, a 2-bit MobileNetV2 trained with BASQ \cite{Kim2022BASQ} uses only 55.3\% as much space as our \methodname 0.75$\times$MobileNetV2, but requires 81.8\% more energy on \SI{7}{\nano\meter} chips for near identical top-1 performance on ImageNet. Hence, there is a trade-off between storage size and energy efficiency, with our proposed method favoring energy-efficient inference over model storage size. 

\subsection{Comparison to Binary Neural Networks}

The concept of low bit-width quantization for convolutional architectures naturally extends to binary networks \cite{Rastegari2016XNOR,Hubara2016Binarized}. Instead of relying on integer addition and multiplication routines, binary networks transform all floating-point operations into logical XNOR and POPCOUNT (counting the number of
ones in a bit-string) commands by quantizing both weights and activations into binary tensors.

We note that neither the authors of these methods \cite{Redfern2021BCNN, Guo2023BNext, Bethge2021MeliusNet, Zhang2022PokeBNN, Liu2020ReActNet, Martinez2020Real2Bin, Rastegari2016XNOR} nor the cost overviews in \cite{Horowitz2014Computings,Zhang2022PokeBNN} provide energy measurements for binary operations. 
Since the energy cost of binary methods cannot be estimated by assuming the use of bit-packing and higher bit-width integer ADD and MUL routines (as we do for int4 and int2 models), we cannot give a fair evaluation to these methods in terms of energy efficiency. Therefore, we limit our comparison with binary networks to model size and architecture design.

\begin{figure}
	\centering
	\includegraphics[width=0.85\linewidth]{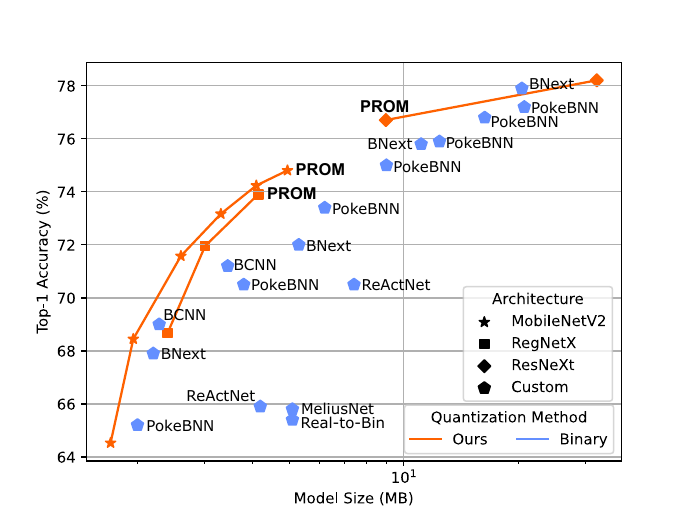}
	\caption{Comparison with binary models in the trade-off between task performance and model size.}
	\label{subfig:memory_pareto_1_bit}
\end{figure}

In terms of the trade-off between model size and top-1 accuracy, we analyze the performance of our method in contrast to binary CNNs in Figure~\ref{subfig:memory_pareto_1_bit}.
Interestingly, our \methodname scheme outperforms most binary models, requiring less model storage space for the same task performance, despite binary methods utilizing 1-bit weights. We attribute this observation to the fact that all tested binary methods use custom model architectures, which might not be as optimized as MobileNetV2 or ResNeXt. This could also be undesirable for commercial use-cases, where optimizations for existing and well-tested models may be preferred.

\subsection{Hardware Support}
As quantization is mostly used to decrease the cost of running a model on a given hardware in terms of memory and energy requirements, hardware support for the proposed quantization operations is essential. Most modern CPU and GPU architectures include highly optimized instructions for int8 operations such as addition and multiplication, rendering our proposed method efficient and widely applicable across a variety of hardware. 

For models employing 2-bit \cite{Gong2019Differentiable, Esser2020LSQ} or 4-bit \cite{Gong2019Differentiable, Esser2020LSQ, Li2021BRECQ, Park2020PROFIT} quantization, an efficient implementation is not straight-forward due to a lack of native hardware support, such as in the popular ARMv8 \cite{armIntroduction}, ARM NEON \cite{ARMNEON} or modern Intel \cite{IntelIntrinsics} instruction sets. While specialized hardware could allow these models to run efficiently, general-purpose hardware often lacks optimizations for operations below int8 \cite{Won2022ULPPACK}. This could lead to complicated memory access patterns, as weights need to be packed and unpacked into larger data types, decreasing the model's efficiency. However, when the hardware support for int4 and int2 operations improves, further reducing the bit-width of the non-ternary components in the network is an interesting direction for future research.
 
In contrast to other quantization methods, binary networks \cite{Redfern2021BCNN, Guo2023BNext, Bethge2021MeliusNet, Zhang2022PokeBNN, Liu2020ReActNet, Martinez2020Real2Bin, Rastegari2016XNOR} rely on logic instructions, performing convolution through bitwise XNOR and POPCOUNT operations. While these instructions can be implemented to a high degree of efficiency, e.g.,\ on FPGAs, most general purpose hardware does not natively support XNOR operations \cite{Ferrarini2022Binary, Zhu2020XOR}, in which case they must be simulated through a combination of XOR and NOT. Furthermore, the POPCOUNT operation often takes multiple clock cycles to execute \cite{Fog2022Instruction} and needs to be chunked for larger register sizes. While the theoretical efficiency gains for binary networks are promising, their practical implementation remains complicated. In contrast, our proposed method relies heavily on efficient int8 addition, which takes only one clock cycle on most modern hardware and benefits massively from large AVX2 or AVX-512 registers, performing up to 64 int8-additions in one clock cycle. 

\subsection{Ablation Study}
\label{sec:ablation_study}
\begin{table}
	\centering
	\caption{Component ablation study for our proposed method. "DW Conv Bit-Width" denotes the bit-width used in depthwise convolutions throughout the model. Similarly, we denote the choice of per-tensor or per-channel quantization for pointwise convolutions as "PW Conv Quantization" }
	\label{tab:ablation}
	\resizebox{1.0\linewidth}{!}{%
		\begin{tabular}{ccccc|ccc}
			\multicolumn{5}{c}{Ablation Settings} & \multicolumn{3}{c}{Top-1 Accuracy (\%)} \\
			\toprule
			\rotatebox{90}{\makecell{DW Conv Bit-Width}} & 
			\rotatebox{90}{\makecell{Cosine Decay Scheduler}} & 
			\rotatebox{90}{\makecell{Weight Decay Reset}} &  
			\rotatebox{90}{\makecell{PReLU Activations}} & 
			\rotatebox{90}{\makecell{PW Conv Quantization}} & 
			\rotatebox{90}{\makecell{1.0$\times$MobileNetV2}} &
			\rotatebox{90}{\makecell{RegNetX-400MF}} & 
			\rotatebox{90}{\makecell{ResNeXt-50(32$\times$4d)}}	\\
			\midrule
			1.58 & \plainxmark & \plainxmark & \plainxmark & Per-Tensor & 53.11 \scalebox{0.7}{$\pm$0.54} & 64.66 \scalebox{0.7}{$\pm$0.75} & 74.68 \scalebox{0.7}{$\pm$0.11} \\
			8 & \plainxmark & \plainxmark & \plainxmark & Per-Tensor & 60.37 \scalebox{0.7}{$\pm$0.39} & 67.12 \scalebox{0.7}{$\pm$0.17} & 76.22 \scalebox{0.7}{$\pm$0.13} \\
			8 & \plaincmark & \plainxmark & \plainxmark & Per-Tensor & 62.83 \scalebox{0.7}{$\pm$1.24} & 66.09 \scalebox{0.7}{$\pm$0.29} & \textbf{76.57} \scalebox{0.7}{$\pm$0.15} \\
			8 & \plaincmark & \plaincmark & \plainxmark & Per-Tensor & 66.88 \scalebox{0.7}{$\pm$0.30} & 67.12 \scalebox{0.7}{$\pm$0.53} & 75.71 \scalebox{0.7}{$\pm$0.04} \\
			8 & \plaincmark & \plaincmark & \plaincmark & Per-Tensor & 68.28 \scalebox{0.7}{$\pm$0.09} & 68.20 \scalebox{0.7}{$\pm$0.48} & 72.82 \scalebox{0.7}{$\pm$0.24} \\
			8 & \plaincmark & \plaincmark & \plaincmark & Per-Channel & \textbf{68.30} \scalebox{0.7}{$\pm$0.21} & \textbf{68.28} \scalebox{0.7}{$\pm$0.47} & 72.85 \scalebox{0.7}{$\pm$0.28}\\
			\bottomrule 
		\end{tabular}
	}
\end{table}
We evaluate the influence of design choices in \methodname by conducting an ablation study in Table~\ref{tab:ablation}. For each setting, we perform three runs using different random seeds, and report the mean resulting accuracies as well as their standard deviation.

We summarize our findings as follows: The usage of a mixed quantization scheme, employing ternary weights for pointwise convolutions and 8-bit weights for depthwise convolutions, is recommended for every architecture, as it offers a comparatively cheap way to increase task performance, both in terms of energy cost and model size. Using a cosine decay learning rate scheduler further improves the quantized model's performance, except for the tested RegNetX variant, which already uses a cosine decay scheduler with linear warmup in its vanilla setting. Resetting the weight decay halfway through training or replacing all ReLU or ReLU6 activations with PReLU are architecture-dependent choices, which require experimentation with the given model. Lastly, quantizing the pointwise convolutions on a per-channel basis rather than per-tensor only offers marginal improvements, which could be explained by variance.

\section{Related Work}
\label{sec:relatedwork}

\textbf{Quantizing Weights and Activations to Integers.} 
A common approach to model compression is integer quantization, which converts 32-bit or 16-bit floating-point weights and activations to lower bit-widths. While 8-bit quantization is widely adpoted for its strong performance and hardware support \cite{Park2020PROFIT,Esser2020LSQ,Zhu2020DSGC}, more aggressive 4-bit or 2-bit schemes often suffer from the reduced model capacity and training instabilities \cite{Park2020PROFIT,Zhu2020DSGC,Gong2019Differentiable}. Various techniques have been proposed to mitigate this, including learnable gradient scaling \cite{Esser2020LSQ}, post-training second-order error analysis \cite{Li2021BRECQ}, multi-stage training \cite{Gong2019Differentiable} and selective layer freezing \cite{Park2020PROFIT}.
Another line of work is learned mixed-precision quantization \cite{Chen2021Towards,Chauhan2023Post,Yang2021FracBits,Wang2020Hardware}, aiming to optimize the bit-widths used in every layer. However, this complicates deployment on general-purpose hardware, as it typically produces a highly fragmented mix of bit-widths.
Recently, binary and ternary quantization methods have demonstrated the ability to eliminate expensive multiplications by reducing the forward pass of a module to simple additions \cite{Wang2023BitNet,Ma2024Era,Zhu2024Scalable}. This principle has enabled efficient large language models like BitNet \cite{Wang2023BitNet} and BitNet b1.58 \cite{Ma2024Era} and has been extended to remove all matrix multiplications from Transformer \cite{Vaswani2017Attention} attention mechanisms \cite{Zhu2024Scalable}.

\textbf{Binary Networks with Logical Operations.} 
Binary neural networks \cite{Rastegari2016XNOR,Hubara2016Binarized,Redfern2021BCNN, Guo2023BNext, Zhang2022PokeBNN,Liu2020ReActNet,Bethge2021MeliusNet,Martinez2020Real2Bin} rely on transforming floating-point operations in a network into logical XNOR and POPCOUNT commands by quantizing both weights and activations into binary tensors. XNOR-Net \cite{Rastegari2016XNOR} and BNNs \cite{Hubara2016Binarized} pioneered the concept of binary CNNs with binary activations and demonstrated their feasibility, but suffered from severe accuracy degradation compared to the floating-point baseline. Plenty of works improved upon this scheme by employing specific training schemes \cite{Liu2018BiReal,Zhang2022PokeBNN}, using knowledge distillation from a real-valued teacher model \cite{Liu2020ReActNet,Martinez2020Real2Bin}, introducing custom block structures or using entirely new model architectures \cite{Redfern2021BCNN, Bethge2021MeliusNet, Zhang2022PokeBNN, Guo2023BNext}. While recent works make considerable progress in closing the gap to floating-point networks, this comes at the cost of increased complexity due to longer and more complex training routines as well as the use of custom architectures. 

\textbf{Model Compression via Pruning.} 
Pruning offers another primary path to model compression \cite{Wimmer2023Dimensionality, He2024Structured}. Unstructured pruning targets individual low-importance weights, ranging from simple magnitude-based removal \cite{Han2016Deep} to more sophisticated methods for identifying redundancy \cite{Wimmer2022Interspace,Wimmer2021COPS}. For practical speedups on general-purpose hardware, structured pruning is used to remove entire filters or channels. These methods can be static, resulting in a fixed architecture \cite{Wen2016Learning, Li2017Pruning}, or dynamic, using gates to toggle components based on input complexity \cite{Gao2019Dynamic, Elkerdawy2022Fire}. Recent work has even demonstrated instant, training-free compression \cite{Meiner2025Data}.
While pruning reduces parameter counts and can be paired with quantization for greater efficiency \cite{Han2016Deep}, exploring this synergy is beyond the scope of this work, but remains a compelling avenue for future research.

\section{Conclusion}
Our proposed method, \nospacemethodname, presents a simple and effective way for ternary quantization of modern depthwise-separable CNNs. In contrast to existing methods which rely on int4 or int2 operations that often lack native hardware support, we perform all computations in the widely-supported and highly optimized int8 format, while retaining competitive performance.
By quantizing the costly pointwise convolutions to ternary weights while keeping activations and low-cost modules in 8-bit, we achieve a favorable trade-off between model accuracy, energy consumption and memory requirements.
We validate our method on the popular ImageNet classification benchmark, where we are able to reduce the energy consumption of tested architectures by more than an order of magnitude while keeping the task performance of the real-valued model. Our method established a highly reproducible and hardware-friendly way to quantize modern CNN architectures, improving the Pareto frontier of efficiency for convolutional models on ImageNet.


\bibliography{references}

\ifarxiv
	\onecolumn 
	
	\twocolumn[
	\begin{center}
		\LARGE\bfseries
		Supplementary Material
		\vspace{2em} %
	\end{center}
	] 
	\appendix
	\section{Distribution of Ternary Weights in Pointwise Convolutions}
\subsection{Distribution at Initialization}
At initialization, the ternary weights in pointwise convolutions exhibit a near uniform distribution between the values -1, 0 and 1, with no notable difference between the layers. This is caused by the initialization scheme used in the MobileNetV2 model, namely He normal initialization \cite{He2015Delving}. For a given weight matrix $W \in \mathbb{R}^{\cout \times \cin \times K \times K}$, the weights are initialized by drawing from a normal distribution:
\begin{equation}
	W_{i,j,k,l} \sim \mathcal{N}\left(0, \frac{2}{\cout}\right) \,.
\end{equation}
To quantize the pointwise convolution to ternary weights, we apply \textit{absmean} quantization. The channel-wise scale factor $\alpha_i$ is computed according to Equation~(\ref{eq:scale_factor_alpha}) from the main text, which is an approximation of the expected value of the weight's absolute value:
\begin{equation}
	\alpha_i \approx \mathbb{E}[|W_i|] \,.
	\label{eq:approx_alpha_with_expectation}
\end{equation}
Since every weight is drawn from a normal distribution, we can compute this expected value:
\begin{equation}
	\mathbb{E}[|W_i|] = \sigma \sqrt{\frac{2}{\pi}} \,,
	\label{eq:expectation_abs_value}
\end{equation}
where $\sigma = \sqrt{2/\cout}$ as above. Note that this value is independent of the chosen output channel $i$. Consequently, when rescaling the weights with $\alpha_i$ before quantization, their variance changes:
\begin{equation}
	\operatorname{Var}\left(\frac{W_{i,j,k,l}}{\alpha_i}\right) = \frac{1}{\alpha_i^2} \operatorname{Var}(W_{i,j,k,l}) 
	\overset{(\ref{eq:approx_alpha_with_expectation}),(\ref{eq:expectation_abs_value})}{\approx} \frac{1}{\sigma^2} \frac{\pi}{2} \sigma^2 = \frac{\pi}{2} \,.
\end{equation}
Now that the variance of the rescaled weight matrix is known, we can derive the distribution of ternary weights after rounding and clamping by observing the amount of weights in between the rounding thresholds $-1/2$ and $1/2$. By integrating the probability density function of the corresponding normal distribution, we get
\begin{equation}
	\mathbb{P}\left(-\frac{1}{2} \leq \frac{W_{i,j,k,l}}{\alpha_i} \leq \frac{1}{2}\right) \approx 0.31 \,,
\end{equation}
so approximately 31\% of weights will be rounded to 0 at initialization. Due to the symmetry of normal distributions, the remaining weights will be rounded and clamped to -1 and 1 in equal parts, with approximately 34.5\% of weights assigned to each value, respectively.

\subsection{Distribution after Training}
\begin{figure*}[hb]
	\centering
	\includegraphics[width=1.0\linewidth]{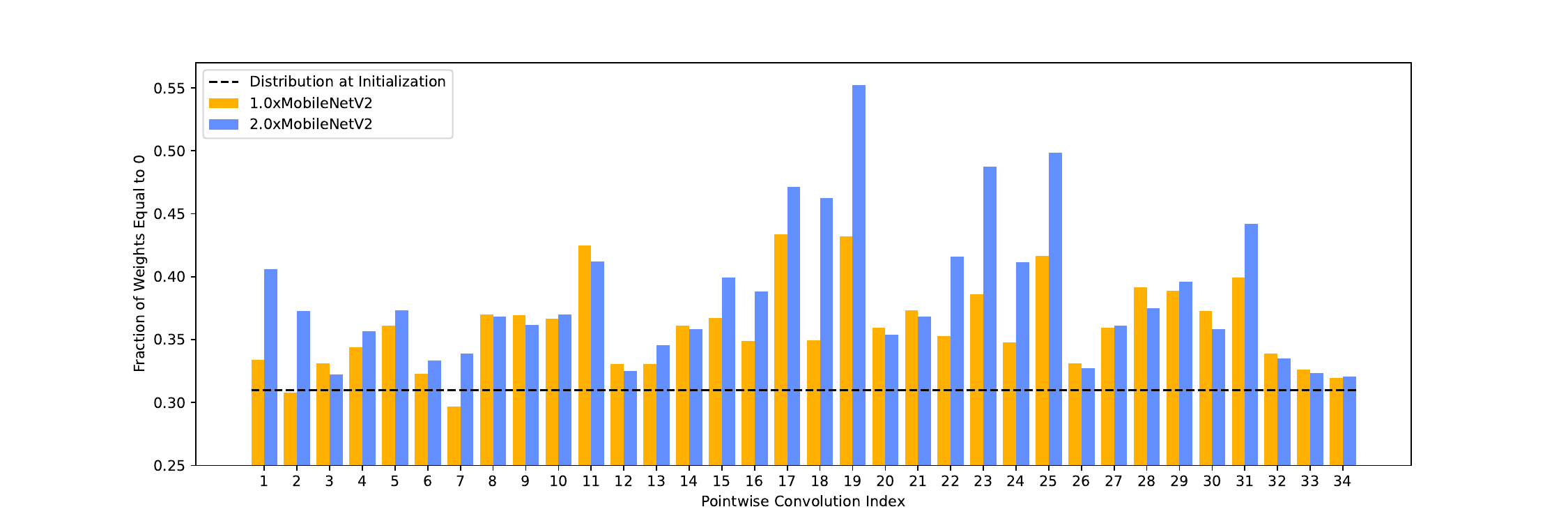}
	\caption{Influence of MobileNetV2's width scaling factor on the distribution of zero weights in the ternary convolutions. \mbox{After} training, the 2.0$\times$MobileNetV2 model contains more weights which are equal to zero.}
	\label{fig:distribution_of_zeros}
\end{figure*}
While the distribution of ternary weights in pointwise convolutions is approximately uniform at initialization, it shifts towards a more uneven one after training, with an increased number of zeros in specific layers. We visualize this finding in Figure~\ref{fig:distribution_of_zeros}. During training, the model seems to automatically learn to "prune" unimportant input connections by setting the corresponding weight to zero. This is particularly noticeable for the 2.0$\times$MobileNetV2. While it uses approximately 3.2$\times$ more parameters than the 1.0$\times$MobileNetV2 model, it does not learn a similar proportion of non-zero weights. Instead, the 2.0$\times$ model exhibits a notably higher percentage of weights equal to zero, with one layer reaching up to 55\% of weights being zero. 

\begin{figure*}
	\centering
	\includegraphics[width=0.7\linewidth]{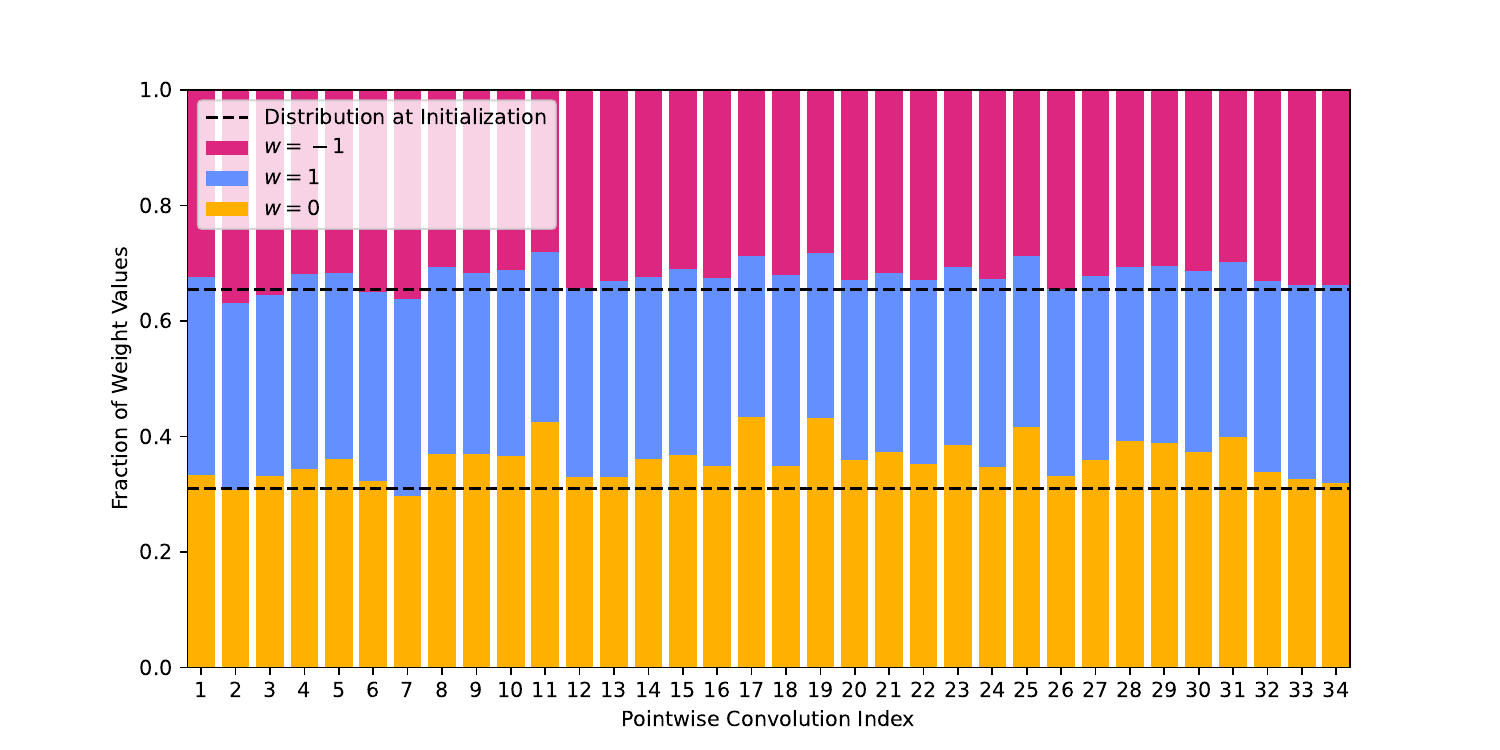}
	\caption{Distribution of ternary weight values in the pointwise convolutions of a $1.0\times$MobileNetV2 after training.}
	\label{fig:distribution_of_ternary_weights_1x}
\end{figure*}

\begin{figure*}
	\centering
	\includegraphics[width=0.7\linewidth]{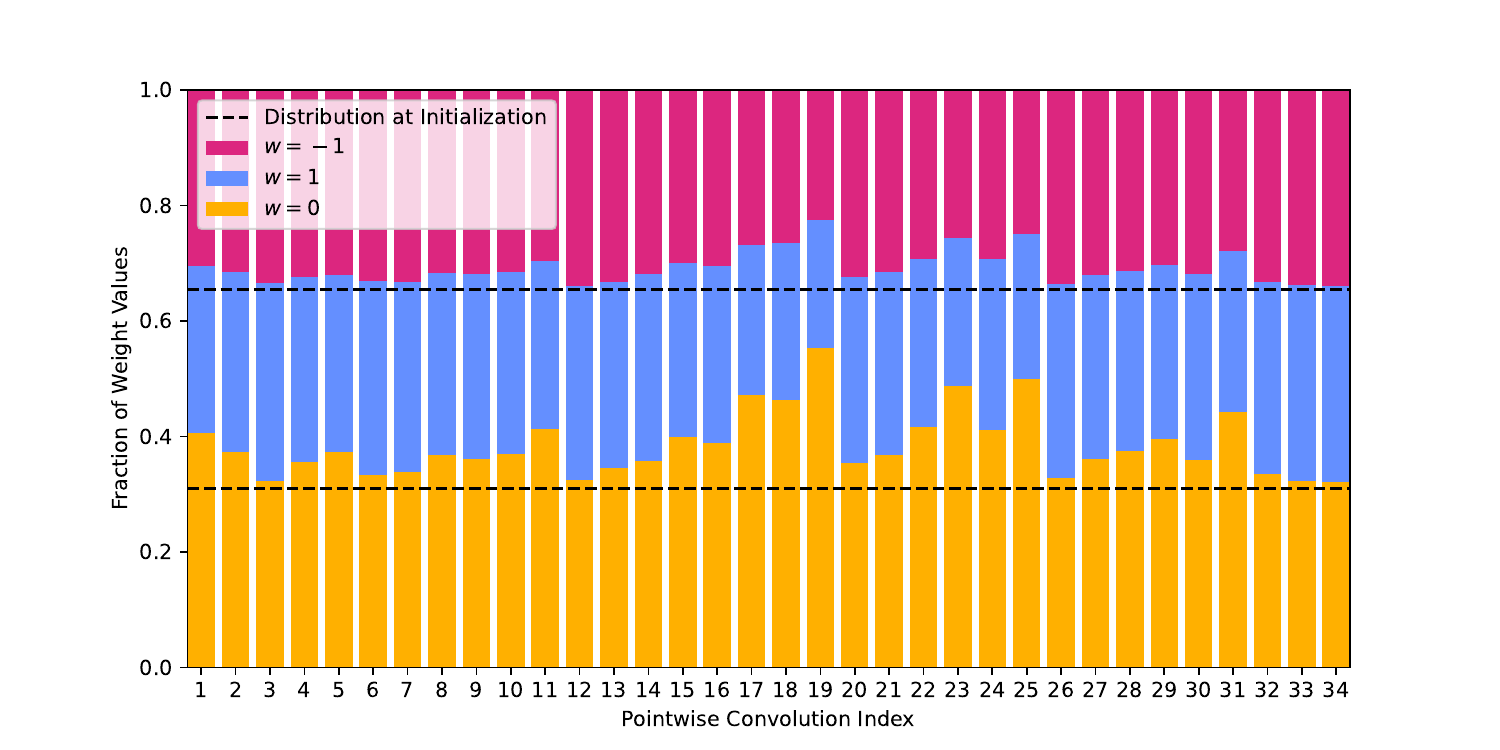}
	\caption{Distribution of ternary weight values in the pointwise convolutions of a $2.0\times$MobileNetV2 after training.}
	\label{fig:distribution_of_ternary_weights_2x}
\end{figure*}

To compare the overall distribution of ternary weights in our smallest and largest model, including -1 and 1, we visualize their layer-wise distribution in Figures~\ref{fig:distribution_of_ternary_weights_1x} and \ref{fig:distribution_of_ternary_weights_2x}.
While the relative amount of zero weights varies throughout both models, we observe that the non-zero weight values are relatively evenly distributed between -1 and 1. This balance of positive and negative weights leads to stable activations with less variability in their magnitude. In part, this behavior may be explained through the usage of BatchNorm \cite{Ioffe2015Batch} directly after pointwise convolutions, which encourages its inputs to be centered, and by initializing the weights in a uniform manner.   

\section{Pseudocode}
We provide pseudocode for our proposed method in the style of PyTorch \cite{Paszke2019PyTorch}. Our method is derived from \cite{Ma2024Era,Ma2024EraSupp} and adapts the ternary quantization scheme to depthwise-separable CNN architectures. 
The pseudocode for the quantization process described in Section~\ref{subsec:quantization_scheme} of the main paper
is presented in Figure~\ref{fig:code_quantization_of_conv_weights}. Pointwise convolutions are quantized to ternary weights using channel-wise \textit{absmean} quantization. Depthwise convolutions are quantized to 8-bit integers using channel-wise \textit{absmax} quantization. 
We also quantize activations to 8-bit integers via tensor-wise \textit{absmax} quantization, as presented in Figure~\ref{fig:code_quantization_of_activations}.
The forward pass of the resulting quantized convolution module is detailed in Figure~\ref{fig:code_quantized_conv_module}.

\begin{figure*}
	\begin{python}
		def quantize_conv(weight, eps = 1e-5):
			"""
			Args:
				weight (Tensor): The weights of the convolution module. 
					Expects weights to have shape [c_out, c_in, k, k].
				eps (float, optional): A small epsilon to prevent division by zero.
			"""
			if weight.shape[2:] == (1,1): # Pointwise convolution
				"""
				Quantize pointwise convolution to ternary weights
				via channel-wise absmean quantization
				"""
				# Compute channel-wise scale factor
				scale = 1.0 / weights.abs().flatten(start_dim=1).mean(dim=-1, keepdim=True).clamp_(min=eps)
				# Reshape the scale factor
				scale = scale.unsqueeze(-1).unsqueeze(-1) # [c_out, 1, 1, 1]
				# Quantize the weights
				quant_weight = (weight * scale).round().clamp_(-1, 1)
				return quant_weight, scale
				
			else: # Depthwise convolution
				"""
				Quantize depthwise convolution to 8-bit weights
				via channel-wise absmax quantization
				"""
				# Compute channel-wise scale factor
				scale = 127.0 / weights.abs().flatten(start_dim=1).max(dim=-1, keepdim=True).values().clamp_(min=eps)
				# Reshape the scale factor
				scale = scale.unsqueeze(-1).unsqueeze(-1) # [c_out, 1, 1, 1]
				# Quantize the weights
				quant_weight = (weight * scale).round().clamp_(-128, 127)
				return quant_weight, scale
	\end{python}
	\caption{Pseudocode for the quantization process of pointwise and depthwise convolution weights.}
	\label{fig:code_quantization_of_conv_weights}
\end{figure*}

\begin{figure*}
	\begin{python}
		"""
		Quantize the activations to 8-bit
		via tensor-wise absmax quantization
		"""
		def quantize_activation(x, eps = 1e-5):
			"""
			Args:
				x (Tensor): The input to be quantized. 
					Expects shape [batch_size, c_in, height, width].
				eps (float, optional): A small epsilon to prevent division by zero.
			"""
			# Compute tensor-wise scale factor
			scale = 127.0 / x.abs().flatten(start_dim=1).max(dim=-1, keepdim=True).values().clamp_(min=eps)
			# Reshape the scale factor
			scale = scale.unsqueeze(-1).unsqueeze(-1) # [batch_size, 1, 1, 1]
			# Quantize the input
			quant_x = (x * scale).round().clamp_(-128, 127)
			return quant_x, scale
	\end{python}
	\caption{Pseudocode for the quantization of activations.}
	\label{fig:code_quantization_of_activations}
\end{figure*}

\begin{figure*}
	\begin{python}
		class QuantizedConv():
			def __init__(self, float_weight):
				""" 
				Args:
					float_weight (Tensor): The underlying (initialized) float weights to train on.
				"""
				self.float_weight = float_weight
				
			def forward(self, x):
				if self.training: # Training pass
					# Quantize the weights on the fly
					quant_weight, scale_weight =  quantize_conv(self.float_weight)
					# Quantize the activation
					quant_x, scale_x = quantize_activation(x)
					
					# Dequantize both before convolving
					quant_weight /= scale_weight
					quant_x /= scale_x
					
					# Straight-through gradient estimator
					quant_weight = self.float_weight + (quant_weight - self.float_weight).detach()
					quant_x = x + (quant_x - x).detach()
				
					output = convolve(quant_x, quant_weight)
					return output		
					
				else: # Inference pass
					# Weights can be quantized and fixed in advance
					quant_weight, scale_weight = quantize_conv(self.float_weight)
					# Quantize the activation
					quant_x, scale_x = quantize_activation(x)		
					
					# Perform convolution in low bit-width
					output = convolve(quant_x, quant_weight)
					
					# Dequantize after convolution
					output /= scale_weight
					output /= scale_x
					
					return output	
					
	\end{python}
	\caption{Pseudocode for a quantized convolution module.}
	\label{fig:code_quantized_conv_module}
\end{figure*}
\fi

\end{document}